\documentclass[sigconf,nonacm]{acmart}

\AtBeginDocument{%
  }

\usepackage{algorithm}
\usepackage{algpseudocodex}
\usepackage{array}
\usepackage{graphicx}
\usepackage{caption}
\usepackage{subcaption}

\begin{document}

\title[GENPACK: KPI-Guided Genetic Algorithm for 3D Bin Packing]{GENPACK: KPI-Guided Multi-Criteria Genetic Algorithm for Industrial 3D Bin Packing}

\author{Dheeraj Poolavaram}
\email{dheeraj.poolavaram@tha.de}
\orcid{0009-0004-5569-0907}
\affiliation{%
  \institution{Technische Hochschule Augsburg}
  \department{TTZ Landsberg am Lech}
  \city{Augsburg}
  \state{Bavaria}
  \country{Germany}
}

\author{Carsten Markgraf}
\email{carsten.markgraf@tha.de}
\affiliation{%
  \institution{Technische Hochschule Augsburg}
  \department{TTZ Landsberg am Lech}
  \city{Augsburg}
  \state{Bavaria}
  \country{Germany}
}

\author{Sebastian Dorn}
\email{sebastian.dorn@tha.de}
\affiliation{%
  \institution{Technische Hochschule Augsburg}
  \department{TTZ Landsberg am Lech}
  \city{Augsburg}
  \state{Bavaria}
  \country{Germany}
}

\renewcommand{\shortauthors}{Poolavaram et al.}

\begin{abstract}

The three-dimensional bin packing problem (3D-BPP) is a longstanding challenge in operations research and logistics. While classical heuristics and constructive methods can generate packings efficiently, they often fail to satisfy industrial requirements such as stability, balance, and handling feasibility. Metaheuristics such as genetic algorithms (GAs) offer greater flexibility, but pure GA approaches frequently struggle with efficiency, parameter sensitivity, and scalability to industrial order sizes. These limitations are particularly evident at real-world pallet dimensions, where even state-of-the-art methods often fail to produce robust, deployable solutions. We propose a KPI-guided GA-based pipeline for industrial 3D-BPP that integrates key performance indicators (KPIs) directly into a scalarized fitness function. The method combines a layer-based chromosome representation, domain-specific operators, and constructive heuristics to balance efficiency and feasibility. On the BED-BPP benchmark of 1,500 real-world orders, our GENPACK pipeline consistently outperforms heuristic and learning-based baselines, achieving up to 35\% higher space utilization and 15--20\% stronger surface support, while exhibiting lower variance across orders. These gains come at a modest runtime cost but remain practical for batch-scale deployment, yielding stable, balanced, and space-efficient packings.

\end{abstract}

\keywords{genetic algorithms, 3D bin packing, multi-criteria optimization, industrial logistics, metaheuristics, chromosome encoding}

\maketitle

\section{Introduction}

The three-dimensional bin packing problem (3D-BPP) is a fundamental NP-hard optimization challenge with direct relevance to logistics, warehouse management, and supply chain operations. The objective is to arrange a set of cuboidal items into a limited number of bins or pallets while minimizing void volume and satisfying geometric feasibility~\cite{martello2000three}. Classical formulations and early heuristics primarily optimize for volume utilization or bin count, but such objectives alone are insufficient for industrial palletizing. In practice, additional requirements such as stability, support, handling feasibility, and weight distribution must also be satisfied for solutions to be deployable~\cite{bortfeldt2013constraints}.

Constructive heuristics such as Bottom-Left~\cite{chazelle1983bottomleft}, Extreme Point~\cite{crainic2008extreme}, and MaxRects~\cite{jylanki2010maxrects} remain widely used because of their efficiency, particularly on large instances~\cite{bischoff1990three}. However, they rely on simplified placement rules that often yield arrangements that fail to satisfy industrial safety and stability constraints. Metaheuristics, and genetic algorithms (GAs) in particular, provide greater flexibility by exploring larger solution spaces and accommodating multiple objectives~\cite{jakobs1996genetic,wu2010three}. However, pure GA approaches also face important limitations: common encodings are not well suited to packing structure, genetic operators often fail to preserve feasibility, and fitness functions frequently do not capture practical indicators of industrial quality~\cite{bortfeldt2001hybrid,egeblad2009heuristic,bortfeldt2013constraints}. Moreover, scalability to real-world order sizes at industrial pallet dimensions, such as Euro-pallets and roll containers, remains a critical challenge where many state-of-the-art methods still struggle~\cite{wu2010three,egeblad2009heuristic}.

Recent work has introduced hybrid methods that combine constructive heuristics with metaheuristic refinement. For example, \mbox{Ananno and Ribeiro}~\cite{ananno2024multiheuristic} proposed a two-stage framework in which constructive heuristics generate layers and blocks, and a GA arranges the remaining residual items under stability and support constraints. While such approaches demonstrate feasibility, they primarily enforce industrial requirements as hard constraints and focus evaluation on compactness and utilization~\cite{bortfeldt2001hybrid,kang2012hybrid,egeblad2009heuristic}. They do not systematically optimize across multiple industrial key performance indicators (KPIs), and runtime scalability remains problematic for highly heterogeneous orders~\cite{egeblad2009heuristic,wu2010three}.

\begin{figure}[!ht]
    \centering
    \includegraphics[width=0.9\linewidth]{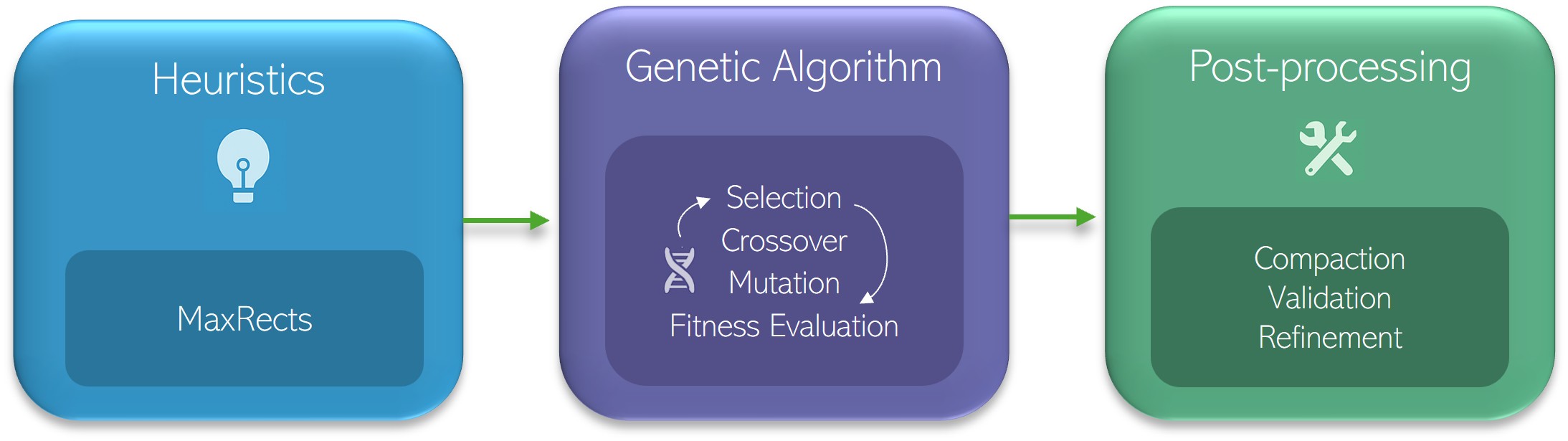}
    \caption{Three-stage pipeline architecture: constructive heuristics generate initial placements, a genetic algorithm refines residuals, and post-processing improves the final configuration.}
    \Description{Diagram showing a three-stage pipeline: constructive heuristics for initial placements, a genetic algorithm for refinement, and post-processing for optimization.}
    \label{fig:architecture}
\end{figure}

This paper addresses these limitations by presenting a KPI-guided GA specifically designed for industrial 3D-BPP. The overall three-stage pipeline of the proposed approach is shown in Figure~\ref{fig:architecture}. In the first stage, constructive heuristics such as MaxRects~\cite{jylanki2010maxrects} rapidly generate initial placements that cover only a subset of the items to be packed. In the second stage, a genetic algorithm uses selection, crossover, mutation, and KPI-based fitness evaluation to refine the packing and place residual items that the heuristics failed to place. In the final stage, post-processing compacts and adjusts the arrangement to improve density and ensure feasibility with respect to stability and industrial constraints.

\noindent The main contributions are as follows:

\begin{itemize}
\item A hybrid pipeline that combines constructive heuristics for initialization with GA-based refinement of residual items to improve scalability.
\item A layer-based chromosome representation that naturally encodes feasible packings while supporting efficient genetic operations.
\item A scalarized KPI-guided fitness formulation integrating industrially relevant KPIs such as density, stability, surface support, balance, and structural integrity.
\item Specialized crossover and mutation operators designed to preserve feasibility and drive targeted KPI improvements.
\item An empirical evaluation on 1,500 real-world orders from the BED-BPP dataset~\cite{kagerer2023bedbpp}, ensuring both reproducibility and comparability.
\end{itemize}

The remainder of the paper is structured as follows. Section~2 reviews related work, Section~3 defines the problem formulation, Section~4 presents the proposed GA methodology, Section~5 describes the hybrid pipeline, Section~6 discusses implementation and dataset characteristics, Section~7 reports experimental results, Section~8 provides a broader discussion, and Section~9 concludes with future directions.

\section{Related Work}

Research on the 3D-BPP spans heuristics, metaheuristics, and learning-based methods, alongside more recent industrially motivated formulations.

\subsection{Heuristics}

Constructive heuristics such as Bottom-Left~\cite{chazelle1983bottomleft}, Extreme Point~\cite{crainic2008extreme}, and MaxRects~\cite{jylanki2010maxrects} generate feasible layouts quickly and scale to large instances~\cite{bischoff1990three}. Their efficiency makes them useful in practice, but they usually rely on simple placement rules that often result in unstable or unbalanced packings, which typically cannot be sufficiently corrected by post-processing to meet industrial constraints.

\subsection{Metaheuristics}

The first application of GAs to packing was introduced by Jakobs~\cite{jakobs1996genetic}, with subsequent work focusing on hybridizing GAs with constructive methods~\cite{wu2010three,kang2012hybrid}. Later studies demonstrated that problem-specific encodings and tailored operators consistently outperform generic representations~\cite{gehring1997genetic,bortfeldt2001hybrid}, and that combining GAs with local search often yields state-of-the-art results~\cite{ceschia2013metaheuristic}. Despite these advances, many GA implementations still rely on simplified fitness functions that fail to capture industrial constraints and continue to face runtime scalability issues, particularly for heterogeneous orders.

\subsection{Learning-based Approaches}

Deep reinforcement learning (DRL) has been applied to online 3D-BPP with promising results. Zhao et al.~\cite{zhao2021constrained} formulated a constrained RL agent with a feasibility mask over a height-map state, improving utilization while suppressing invalid actions during training. Zhao et al.~\cite{zhao2022pct} subsequently introduced packing configuration trees (PCT), a hierarchical representation that enumerates candidate placements and supports efficient policy learning in continuous spaces. Que et al.~\cite{que2023transformer} proposed a Transformer-based PPO policy with a redesigned state and action factorization, and reported strong utilization on synthetic benchmarks. More recently, Xiong et al.~\cite{xiong2024gopt} proposed GOPT, a Transformer-based DRL framework that combines a Placement Generator and a Packing Transformer to stabilize the action space and capture spatial relations, achieving strong generalization across varying bin sizes and unseen items.

Learning-based methods replace explicit search with learned inference and deliver fast test-time decisions, but they often require large training corpora and careful reward and constraint design, and they can be sensitive to domain shift between synthetic training distributions and industrial orders. Adapting them to specific operational rules or KPI trade-offs often necessitates retraining, and additional post-processing may still be required for feasibility on out-of-distribution cases~\cite{zhao2021constrained,zhao2022pct,que2023transformer,xiong2024gopt}.

\subsection{Industrial Constraints}

Beyond maximizing utilization, research has increasingly emphasized constraint-aware optimization and load balancing~\cite{bortfeldt2013constraints,ramos2018new,fanslau2010tree}. Surveys indicate that practical requirements such as accessibility, handling feasibility, and stability remain insufficiently addressed in optimization objectives~\cite{pantoja2024comprehensive,wu2023machine}. Some recent frameworks~\cite{ananno2024multiheuristic} incorporate these industrial aspects primarily as hard constraints while still prioritizing utilization and compactness. However, when applied at real-world pallet scale, such approaches often encounter significant runtime limitations.

\subsection{Benchmarking}

Traditional benchmarking has largely relied on synthetic datasets such as the Bischoff and Marriott instances~\cite{bischoff1990three} and, more recently, software toolboxes for dataset generation~\cite{ribeiro2023toolbox}. While useful for reproducibility, these datasets are either fully synthetic or designed primarily for multi-container problems, and thus fail to capture the heterogeneity and operational constraints of single-container, real-world palletization scenarios. In contrast, the BED-BPP dataset~\cite{kagerer2023bedbpp}, with more than 10{,}000 grocery orders including precise item dimensions and corresponding weights, enables realistic, industry-grounded evaluation and has therefore been adopted for empirical analysis in our work.

The above studies demonstrate the efficiency of heuristics in generating fast and interpretable packing solutions, the flexibility of metaheuristics in exploring diverse search spaces and refining constructive outputs, and the promise of learning-based policies in capturing structural patterns. At the same time, they also reveal limitations, such as simplified objective formulations that overlook practical KPIs like stability, load balance, or number of pallets, as well as runtime bottlenecks when scaling to real-world pallet dimensions. These gaps collectively motivate the design of a KPI-guided hybrid genetic algorithm in this work, aiming to maximize computational efficiency while maintaining practical relevance at industrial scale.

\section{Problem Formulation}

\subsection{Setting and Variables}

Given is a set of items $\mathcal{I}=\{1,\dots,n\}$, where item $i$ has dimensions $(l_i,w_i,h_i)$ and mass $m_i$. Bins (pallets) share base $(L,W)$, height $H$, and payload capacity $M_{max}$. Items are assumed to be axis-aligned in a right-handed coordinate system where $z$ is normal to the pallet base. Rotation is restricted to $90^\circ$ multiples around the $z$-axis, which allows interchanging the $x$- and $y$-dimensions of an item while keeping its height upright. This reflects common industrial requirements, where most packaged goods can be safely turned sideways on the pallet but should not be tilted onto their ends, as such orientations risk compromising structural integrity or violating handling conventions~\cite{bortfeldt2013constraints}. A packing solution assigns each item to one bin with a placement $(x_i,y_i,z_i)$, denoting the coordinates of its lower-left-front corner, i.e., the corner at minimal $(x,y,z)$ in the global reference frame, together with a rotation $\theta_i$.

\subsection{Feasibility}
\label{subsec:feasibility}

A packing solution is considered feasible when it satisfies the following conditions:  
(i) \textit{assignment} --- each item is placed in exactly one bin;  
(ii) \textit{containment} --- all items lie fully within the bin boundaries;  
(iii) \textit{non-overlap} --- no items in the same bin intersect; and  
(iv) \textit{payload} --- the total mass of items in each bin does not exceed $M$.  

These constraints are enforced during both construction and refinement, for example by axis-disjunctive checks to prevent overlaps and per-bin mass accumulation to enforce payload limits~\cite{martello2000three}. Additional rules, such as minimum support or restricted orientations, can be activated depending on the specific problem instance~\cite{bortfeldt2013constraints}.

\subsection{Objective: Multi-KPI Optimization}

Classical formulations of bin packing primarily aim to minimize the number of bins or maximize space utilization~\cite{martello2000three}. In industrial palletizing, however, additional aspects such as stability, surface support, balance, and handling feasibility are equally critical~\cite{bortfeldt2013constraints}. To account for these practical requirements, we formulate the optimization as a weighted multi-criteria problem over a set of $K$ key performance indicators (KPIs):
\begin{equation}
C^{*} = \arg\max_{C}
\big(\alpha_{1}\,\mathrm{KPI}_{1}(C) + \alpha_{2}\,\mathrm{KPI}_{2}(C) + \dots + \alpha_{K}\,\mathrm{KPI}_{K}(C)\big),
\end{equation}

\noindent where $C$ denotes a feasible packing configuration, represented by a set of layers $\mathcal{L}_i$ with assigned item placements and orientations (collectively forming the chromosome $C$ in~\eqref{eq:chromosome}). In this work, the KPI set comprises density, stability, surface support, balance, and structural integrity metrics. For integration within the genetic algorithm (Section~\ref{sec:method_ga}), this multi-criteria formulation is scalarized into a single aggregated fitness score.

\newcommand{\kpi}[1]{\text{\textsc{#1}}}
\newcommand{\Overlap}{\operatorname{Overlap}}

\section{Genetic Algorithm Approach}
\label{sec:method_ga}

\subsection{Chromosome Representation}

Each chromosome encodes a packing as an ordered list of layers
\begin{equation}
C = [\mathcal{L}_1, \mathcal{L}_2, \dots, \mathcal{L}_m],
\label{eq:chromosome}
\end{equation}

\noindent where each layer $\mathcal{L}_j$ stores (i) its assigned items, (ii) their 2D coordinates $\{(x_i,y_i)\}$, (iii) its height $h(\mathcal{L}_j)$, i.e., the maximum height among contained items, and (iv) the layer's base $z$-level. Layers are stacked sequentially, and the cumulative height defines the overall packing height. Within a layer, items are axis-aligned and placed on a coarse grid that avoids 2D overlaps, while vertical overlaps are prevented by stacking order.

\subsection{Multi-KPI Fitness Function}
\label{subsec:kpis}

A single scalar fitness aggregates multiple KPIs (seven in this work) and an overlap penalty. For chromosome $C$:
\begin{equation}
\label{eq:fitness}
\mathrm{Fitness}(C)
= \sum_{k=1}^{7} \alpha_{k}\,\mathrm{KPI}_k(C) - \min\!\big(1,\,\Overlap(C)\big),
\end{equation}
where
\begin{equation*}
\begin{aligned}
\{\mathrm{KPI}_k\}_k = \{&
\textsc{Coverage}, \textsc{AbsDen}, \textsc{RelDen}, \textsc{SideSup},\\
&\textsc{SurfSup}, \textsc{TallItem}, \textsc{CoG2D}\}.
\end{aligned}
\end{equation*}

\noindent and each $\mathrm{KPI}_k(C)\in[0,1]$, with 1 denoting the optimum. The weights $\alpha_{k}$ are non-negative and sum to one. They were chosen from ablation and sensitivity studies, aligned with practical palletizing priorities, and can be adjusted for different use cases. The overlap fraction is normalized by the bin volume:
\begin{equation}
\Overlap(C) \;=\; \frac{1}{LWH}\sum_{\substack{i\neq j}} \mathrm{VolOverlap}(i,j),
\label{eq:overlap}
\end{equation}
where $\mathrm{VolOverlap}(i,j)$ is the 3D intersection volume between items $i$ and $j$ (zero if none). Let $P(C)$ denote the set of items that are successfully placed in configuration $C$. We define the used stacking height as:
\begin{equation}
H^\ast \;=\;
\begin{cases}
\max\limits_{i\in P(C)}\big(z_i + h_i\big), & \text{if } P(C)\neq \varnothing,\\[4pt]
0, & \text{if } P(C)=\varnothing.
\end{cases}
\label{eq:used-height}
\end{equation}

\noindent The KPIs in Eq.~\eqref{eq:fitness} combine standard measures of volume utilization, support, and center-of-gravity balance, which are widely used in container loading and 3D-BPP studies~\cite{martello2000three,bortfeldt2013constraints,crainic2008extreme,ramos2018new,elhedhli2019three}, with problem-specific extensions (e.g., \kpi{RelDen}, \kpi{TallItem}) that we introduce to capture compactness and stability in pallet loading more effectively. They are defined as follows:

\begin{itemize}
    \item \textsc{Coverage}: share of unique residual items in $R$ (those not placed by the heuristic phase) that are successfully packed into $C$; each item is counted once~\cite{martello2000three,bischoff1990three}. 

    \item \textsc{AbsDen}: absolute density within the used height, a refinement of the standard volume-utilization measure widely used in bin packing and container loading~\cite{martello2000three,crainic2008extreme}.
    \begin{equation}
    \kpi{AbsDen}(C) = \frac{\sum_{i\in P(C)} \mathrm{vol}(i)}{L\,W\,H^\ast}.
    \end{equation}

    \item \textsc{RelDen}: relative density below the upper envelope, capturing void reduction and local compactness. While compaction is a common heuristic principle in 3D-BPP algorithms~\cite{crainic2008extreme,egeblad2009heuristic}, this envelope-based voxel KPI is specific to this work. Using a voxel grid of spatial resolution $r$ discretizing the pallet base $(L,W)$ and height $H^\ast$, let $\mathcal{G}[x,y,z]\in\{0,1\}$ denote occupancy (1 if a voxel is filled). Define the upper envelope
    \begin{equation}
    h_{upp}(x,y) = \max\{\, z : \mathcal{G}[x,y,z]=1 \,\}\,,
    \end{equation}
    the set $\mathcal{U}=\{(x,y,z)\,:\, 0\le z < h_{upp}(x,y)\}$ of voxels lying strictly below the envelope, and the subset $\mathcal{O}=\{(x,y,z)\in\mathcal{U}\,:\,\mathcal{G}[x,y,z]=1\}$ of those voxels that are occupied. Then
    \begin{equation}
    \kpi{RelDen}(C) = \frac{|\mathcal{O}|}{|\mathcal{U}|},  
    \end{equation}
    which measures the fraction of voxels below the envelope that are filled.

    \item \textsc{SideSup}: lateral support. Stability requirements that restrict unsupported side faces are common in container and pallet loading~\cite{bortfeldt2013constraints}. For each item, we compute the area fraction of each vertical face that is in contact with neighboring items or pallet walls; a face is considered supported if at least $20\%$ of its area is covered. \textsc{SideSup} is the average supported-face rate over all items.

    \begin{equation}
    \textsc{SideSup}(C)
    = \frac{1}{|\mathcal{P}|}
      \sum_{i\in\mathcal{I}}
      \frac{1}{|\mathcal{B}_i'|}
      \sum_{f\in\mathcal{B}_i'}
      \mathbf{1}\!\left(
        s_{i,f}\ge\tau_{\mathrm{side}}
      \right),
    \label{eq:sidesup}
    \end{equation}
    where
    $s_{i,f}=A_{i,f}^{\mathrm{contact}}/A_{i,f}$ 
    is the fraction of face area in contact with other items,
    $\tau_{\mathrm{side}}=0.2$ is the support threshold,
    $\mathcal{B}_i'$ is the set of non-boundary vertical faces of item~$i$,
    and $\mathcal{P}$ is the set of all packed items.

    \item \textsc{SurfSup}: vertical support. Minimum base-support requirements are standard in pallet loading and stacking stability~\cite{bortfeldt2013constraints,elhedhli2019three}. We measure the fraction of non-ground items whose bottom-face overlap exceeds $75\%$.

    \item \textsc{TallItem}: penalty-based stability for tall or slender items, motivated by the instability of tower-like or off-center stacks discussed in container-loading stability studies~\cite{bortfeldt2013constraints}. The score penalizes large height-to-base ratios and off-center or elevated placements.

    For item $i$, define the height-to-base ratio
    \begin{equation}
    r_i =\frac{h_i}{\sqrt{w_i\,l_i}}\,,
    \end{equation}
    and let $d_{\max}=\sqrt{(L/2)^2+(W/2)^2}$. For items with $r_i>1$, let $d_i$ be the Euclidean distance from the footprint center to $(L/2,W/2)$ and define the per-item penalty
    \begin{equation}
    p_i = r_i \left(0.7\,\frac{d_i}{d_{\max}} \;+\; 0.3\,\frac{z_i}{H^\ast}\right).
    \end{equation}
    The score averages these penalties over tall items and maps to $[0,1]$:
    \begin{equation}
    \kpi{TallItem}(C) = \max\!\left(0, 1 - \frac{1}{|\{i:\,r_i>1\}|}\sum_{i:\,r_i>1} p_i \right).
    \end{equation}
    This rewards placing tall items closer to the center and lower in the stack.

    \item \textsc{CoG2D}: balance in the plane. CoG placement is a widely used stability criterion in palletization and load-balancing literature~\cite{ramos2018new,elhedhli2019three}. Let $(x_c,y_c)$ be the total 2D center of gravity projected onto the pallet plane, and let
    \begin{equation}
    \delta = \frac{\sqrt{(x_c-L/2)^2 + (y_c-W/2)^2}}{d_{\max}},
    \end{equation}
    where $\delta$ denotes the normalized radial distance (Euclidean offset) of the total 2D center of gravity from the pallet center, and
    \begin{equation}
    \kpi{CoG2D}(C) = 1 - \delta.
    \end{equation}
\end{itemize}

\noindent All KPI thresholds (e.g., the $20\%$ side-contact threshold in \textsc{SideSup} and the $75\%$ base-overlap threshold in \textsc{SurfSup}) were empirically determined through ablation studies and can be adjusted for different problem settings. They are consistent with established practices in container and pallet loading~\cite{bortfeldt2013constraints}.

\subsection{Population Initialization}

The initial population $\mathbf{P}_0$ is built from residual superitems $R$ that were not included in the constructive phase solution $S_{0}$, i.e., items left unpacked during the heuristic phase. It follows the procedure in Algorithm~\ref{alg:kpi-ga}, combining multiple low-cost heuristics to ensure diversity. Specifically, items are sorted by volume (\textsc{VolSort}), by base area (\textsc{AreaSort}), or left in random order (\textsc{Random}); spatial placement is then performed layer-wise on a coarse grid using the \textsc{BottomLeft} rule~\cite{chazelle1983bottomleft,crainic2008extreme} or an \textsc{ExtremePoint} (\textsc{EP}) strategy~\cite{crainic2008extreme}.

\subsection{Evolutionary Operators}

\subsubsection{Selection and Replacement}

We use tournament selection with adaptive pressure defined as
\begin{equation}
\max\{3,\lfloor \text{POP}(0.1 + 0.3\,g/G)\rfloor\},
\end{equation}

\noindent where POP denotes the population size, $g$ the current generation, and $G$ the total number of generations. This formulation gradually increases the tournament size from roughly 10\% to 40\% of the population. The best 10\% of individuals (elitism) are preserved in each generation, and the remaining offspring are produced from the selected parents.

\subsubsection{Crossover and Mutation}

A one-point \textit{crossover} with layer preservation is applied with probability
\begin{equation}
P_c=0.5(1+0.1\,g/G),
\end{equation}
exchanging layer suffixes between parents and recomputing $z$-levels and heights. If feasibility cannot be maintained, the parents are copied unchanged.

\noindent \textit{Mutation} occurs with probability
\begin{equation}
P_m = 0.35(1 - 0.5\,g/G),
\end{equation}
favoring exploration early in evolution. Each mutation operator $\text{MUT}_k \in \{\text{MUT}_k\}_k$ corresponds to a KPI-aligned adjustment. Specifically, mutations include shifting tall items toward the center or lower layers to improve \textsc{TallItem} and \textsc{CoG2D}, filling small 2D voids below items to improve \textsc{RelDen}, compacting arrangements toward the origin to enhance \textsc{AbsDen} and \textsc{RelDen}, optimizing lateral contact to improve \textsc{SideSup}, and increasing bottom overlap to strengthen \textsc{SurfSup}. Additional diversification operations such as item, layer, or layer-addition swaps are also applied.

\subsection{Parameters}

We use a population size of 100, up to 50 generations, an elite rate of 10\%, and adaptive $P_c$ and $P_m$ for crossover and mutation, respectively, as described above. Fitness values are clipped to a small positive lower bound for numerical stability. All parameter values fall within ranges commonly adopted in GA-based three-dimensional bin packing and container loading literature~\cite{jakobs1996genetic,wu2010three,kang2012hybrid,bortfeldt2013constraints,ananno2024multiheuristic}. Sensitivity experiments within typical ranges ($P_c\in[0.4,0.6]$, $P_m\in[0.2,0.4]$, population $=50$--$200$) confirmed similar convergence behavior. The adaptive scheduling of $P_c$ and $P_m$ gradually increases exploitation and reduces randomness toward later generations, a standard mechanism for stabilizing GA search~\cite{eiben1999parameter}.

The best chromosome is then converted back to the global solution by aligning its stacked layers with the base $z$-level from the constructive phase solution $S_{0}$. The overall GENPACK optimization loop is outlined in Algorithm~\ref{alg:kpi-ga}.

\begin{algorithm}[t]
\caption{GENPACK: Hybrid KPI-Driven GA for Residual 3D-BPP}
\label{alg:kpi-ga}
\begin{algorithmic}[1]
\Require Residual superitems $R$, base solution $S_0$, pallet $(L,W,H)$, mass $m$, pop.~size $P$, generations $G$
\State $\mathbf{P}_0 \gets \textsc{InitializePopulation}(R,$
\Statex \hspace{\algorithmicindent}
$\{\textsc{VolSort}, \textsc{AreaSort}, \textsc{BottomLeft}, \textsc{EP}, \textsc{Random}\}, P)$
\ForAll{$C \in \mathbf{P}_0$}
    \State $f(C)\gets \textsc{EvaluateFitness}(C)$
\EndFor
\State $S_{GA} \gets \arg\max_{C\in \mathbf{P}_0} f(C)$;\quad $\mathbf{P} \gets \mathbf{P}_0$
\For{$g \gets 1$ \textbf{to} $G$}
  \State $t \gets \max\{3,\lfloor P(0.1+0.3\,g/G)\rfloor\}$
  \State $P_c \gets 0.5(1+0.1\,g/G)$;\quad $P_m \gets 0.35(1-0.5\,g/G)$
  \State $\mathbf{E} \gets$ top $\lceil 0.10P\rceil$
  \State $\mathbf{P}_{\text{new}} \gets \mathbf{E}$
  \While{$|\mathbf{P}_{\text{new}}| < P$}
    \State $p_1,p_2 \gets \textsc{TournamentSelect}(\mathbf{P},t)$
    \If{$\textsc{Rand}()<P_c$}
        \State $c \gets \textsc{Crossover}(p_1,p_2)$
    \Else
        \State $c \gets \textsc{Copy}(\arg\max\{f(p_1),f(p_2)\})$
    \EndIf
    \If{$\textsc{Rand}()<P_m$}
      \State Apply $\text{MUT}_k \in \{\text{MUT}_k\}_k$ on $c$
    \EndIf
    \State $f(c) \gets \textsc{EvaluateFitness}(c)$
    \State $\mathbf{P}_{\text{new}} \gets \mathbf{P}_{\text{new}} \cup \{c\}$
  \EndWhile
  \State $\mathbf{P} \gets \mathbf{P}_{\text{new}}$; update $S_{GA}$
\EndFor
\State $S^\star \gets \textsc{MergeWithBase}(S_0,S_{GA})$
\State \Return $S^\star$
\end{algorithmic}
\end{algorithm}

\begin{algorithm}[t]
\caption{Post-Processing via Directional Compaction, Fallback, and Validation}
\label{alg:postprocessing}
\begin{algorithmic}[1]
\Require GA output $S^\star$, pallet $(L,W,H)$, support threshold $\tau_s$
\Procedure{PostProcess}{$S^\star, (L,W,H), \tau_s$}
  \State $\mathcal{L} \gets \text{layers}(S^\star)$
  \ForAll{$\mathcal{L}_j \in \mathcal{L}$}
    \State Sort items by $x_i+y_i$ (proximity to origin)
    \Repeat
      \ForAll{$i \in \mathcal{L}_j$}
        \State Propose local shift $(x_i',y_i')$ toward origin
        \If{feasible and $\rho_{\text{support}}(i)\ge\tau_s$}
          \State Accept if $(x_i'+y_i')<(x_i+y_i)$
        \EndIf
      \EndFor
    \Until{no item improves}
  \EndFor
  \Statex
  \ForAll{unplaced items $i$}
    \State Evaluate candidate $(x,y,z)$ near corners and edges
    \If{none valid} \State perform coarse grid search with rotation \EndIf
    \If{first feasible position found} \State place $i$ \EndIf
  \EndFor
  \Statex
  \ForAll{items $i$ in $\mathcal{L}$}
    \If{overlaps or $\rho_{\text{support}}(i)<\tau_s$}
      \State remove $i$ from packing
    \EndIf
  \EndFor
  \State \Return $S^{\star}_{\text{pp}} \gets \text{validated packing from }\mathcal{L}$
\EndProcedure
\end{algorithmic}
\end{algorithm}

\section{Hybrid Pipeline Architecture}

This section outlines the overall workflow that combines constructive heuristics, genetic refinement, and post-processing into a unified pipeline. Our method follows a three-stage placement strategy: in the first stage, constructive heuristics generate an initial packing using MaxRects; in the second stage, a genetic algorithm refines the residual items missed by MaxRects through KPI-weighted optimization; and in the final stage, the refined solution is merged back into the full configuration and post-processed.

\subsection{Superitem Preprocessing}

To reduce the effective search space, similar items are grouped into superitems for the heuristics~\cite{elhedhli2019three}. Three categories are considered: single superitems represent individual items, horizontal superitems combine multiple items of identical height, and vertical superitems stack compatible items into stable columns up to a maximum height. These groupings reflect practical packing patterns and improve scalability by reducing the number of decision variables while preserving feasible packing opportunities.

\subsection{Integration Strategy}

After GA optimization, residual packings are merged with the constructive baseline (Section~\ref{sec:method_ga}). Layers are aligned with the stacking height produced in the first stage, and all placements are validated to ensure feasibility (Section~\ref{subsec:feasibility}); KPI scores are then recomputed for the combined solution.

\subsection{Post-Processing}

Although GA refinement improves density, stability, and balance, it does not always ensure full feasibility, as defined in Section~\ref{subsec:feasibility}. A small fraction of items may remain unplaced, or minor density irregularities may weaken overall packing quality. Post-processing addresses these issues through a multi-stage heuristic procedure (Algorithm~\ref{alg:postprocessing}). The mathematical formulation of the post-processing is discussed in detail in Appendix~\ref{app:postprocessing}.

The procedure begins with directional compaction, where items in each layer are iteratively shifted toward the pallet origin to minimize empty space while preserving support stability. In the second stage, fallback placement handles any unplaced items by evaluating feasible positions near pallet boundaries and, if necessary, performing a coarse grid search with rotation to recover remaining items. Finally, a validation stage ensures full feasibility by removing overlapping items or those failing the minimum support threshold.

Empirically, this step may slightly alter certain KPIs, particularly side support, but it significantly strengthens surface support, which is the primary indicator of physical stability in real-world pallet configurations~\cite{bortfeldt2013constraints}. Overall, post-processing transforms GA-optimized packings into fully feasible, industrially compliant solutions with improved structural integrity.

\subsection{Computational Complexity}

Let $n$ be the number of items, $P$ the population size, and $G$ the number of generations. In our implementation, fitness evaluation dominates runtime. Each call to \texttt{fitness\_function} computes side support, surface support, and an explicit overlap penalty using pairwise item checks, yielding $\Theta(n^2)$ work per chromosome. Here $\Theta(n^2)$ denotes a tight asymptotic bound, i.e., both the upper and lower order of growth are quadratic in the number of items. A fixed $20\times20\times20$ voxel grid is used for density/holes; this contributes only a constant overhead per evaluation relative to $n$.

Per generation, we therefore spend $\Theta(P\,n^2)$ on fitness, plus $O(P\log P)$ for elite selection and up to $O(P^2)$ from tournament selection in reproduction. For practical settings where $n$ is moderate to large, the overall cost scales as
\[
T(G) \;=\; \Theta\big(G\,(P\,n^2)\big)\quad\text{(fitness-dominated)}.
\]
Population storage is $\Theta(P\,n)$ (chromosomes store item references and coordinates), while the voxel grid used inside fitness is transient. While the quadratic cost per chromosome may appear expensive, it is in fact typical for 3D packing algorithms, since explicit overlap detection requires comparing each item against all others in the same bin~\cite{martello2000three,wu2010three,bortfeldt2013constraints,kang2012hybrid}. Given pallet-scale problem sizes (hundreds of items), this cost is manageable.

\section{Experimental Design}

\subsection{Dataset Characteristics}

Experiments were conducted on a subset of orders from the BED-BPP dataset~\cite{kagerer2023bedbpp}, which contains over 10{,}000 real-world grocery orders. The dataset was tailored to Euro-pallets with a $1200 \times 800$\,mm base. Orders were filtered to ensure realistic palletizing conditions: either at least five distinct item footprints with an average stacking height between 150--190\,cm, or at least nine footprints with average height between 120--150\,cm. For evaluation, we used a balanced subset of 1{,}500 orders that preserves the distribution of dimensions, weights, and product categories, providing a representative benchmark for heterogeneous pallet packing tasks.

\subsection{Evaluation Methodology}

Performance was evaluated in terms of solution quality and computational efficiency. The GA optimizes seven KPIs (Section~\ref{subsec:kpis}), but the evaluation uses a slightly broader set: \textsc{AbsDen, RelDen, H/W, SideSup, SurfSup, CoG2D, and CoG3D.} The height--width ratio \textsc{(H/W)} is a simple compactness measure defined as the used stacking height divided by the pallet width; it is used only for evaluation and is not part of the GA fitness. \textsc{Coverage} and \textsc{TallItem} are likewise fitness-only objectives. To ensure fair comparison, all KPI values were normalized by the corrected packing efficiency (valid\_items / total\_items), penalizing infeasible packings where any item violates spatial constraints. Runtime was used to assess efficiency, and convergence and robustness were examined through fitness evolution across generations and heterogeneous dataset subsets.

\subsection{Baselines}

We compared GENPACK against seven algorithms covering classical heuristics, learning-based methods, and our constructive--hybrid variants (Table~\ref{tab:baselines}). External baselines include O3D-BPP-PCT~\cite{zhao2022pct}, GOPT~\cite{xiong2024gopt}, Extreme Point~\cite{crainic2008extreme}, and Sisyphus~\cite{demisse2012mixed}. Internal baselines include MaxRects~\cite{jylanki2010maxrects}, GENPACK, and GENPACK (+PP), which represent successive stages of our pipeline and isolate the effect of each component.

This selection ensures balanced evaluation across three categories: classical heuristics that remain widely used in practice, recent learning-based approaches, and ablated hybrid variants showing the contributions of evolutionary refinement and post-processing.

All experiments were conducted on a machine with an Intel Core Ultra 7\,155U processor (1.70\,GHz), 32\,GB RAM, and Windows\,11\,Pro. The full implementation is available at \href{https://github.com/dheerajpr97/genpack-3d-bpp-anonymous}{GitHub} for reproducibility.

\begin{table*}[t]
\centering
\caption{Overview of baseline algorithms and their categories.}
\label{tab:baselines}
\begin{tabular}{llp{11cm}}

\toprule
\textbf{Algorithm} & \textbf{Category} & \textbf{Description} \\
\midrule
O3D-BPP-PCT~\cite{zhao2022pct} & Learning-based & Reinforcement learning method using packing configuration trees. \\
GOPT~\cite{xiong2024gopt} & Learning-based & Deep reinforcement learning method using a Transformer architecture. \\
Sisyphus~\cite{demisse2012mixed} & Heuristic & Industrially motivated constructive rules using handcrafted heuristic strategies. \\
Extreme Point~\cite{crainic2008extreme} & Heuristic & Constructive heuristic based on extreme-point placement rules. \\
MaxRects~\cite{jylanki2010maxrects} & Heuristic & Constructive baseline using the MaxRects placement strategy.\\
GENPACK (core) & Hybrid (ours, ablation) & Extends MaxRects with GA refinement of residuals but without post-processing. \\
GENPACK (+PP) & Hybrid (ours, final) & Full proposed pipeline: MaxRects, GA refinement, and post-processing.\\
\bottomrule
\end{tabular}
\end{table*}

\begin{table*}[t]
\centering
\caption{KPIs for packing efficiency and stability (mean $\pm$ std).}
\label{tab:kpi-density}
\begin{tabular}{lccc}
\toprule
Algorithm & AbsDens & RelDens & H/W \\
\midrule
O3D-BPP-PCT & 0.344 $\pm$ 0.183 & 0.462 $\pm$ 0.138 & 0.535 $\pm$ 0.127 \\
GOPT        & 0.360 $\pm$ 0.127 & 0.330 $\pm$ 0.123 & 0.516 $\pm$ 0.122 \\
Sisyphus    & 0.319 $\pm$ 0.097 & 0.455 $\pm$ \underline{0.086} & 0.530 $\pm$ \underline{0.063} \\
Extreme Point & 0.453 $\pm$ \underline{0.092} & 0.485 $\pm$ 0.103 & 0.611 $\pm$ 0.102 \\
MaxRects    & 0.239 $\pm$ 0.104 & 0.466 $\pm$ 0.099 & 0.442 $\pm$ 0.092 \\
GENPACK (core) & \textbf{0.538} $\pm$ \textbf{0.082} & \textbf{0.657} $\pm$ \textbf{0.072} & \textbf{0.696} $\pm$ \textbf{0.049} \\
GENPACK (+PP) & \underline{0.469} $\pm$ 0.098 & \underline{0.531} $\pm$ 0.119 & \underline{0.651} $\pm$ 0.087 \\
\bottomrule
\end{tabular}
\end{table*}

\begin{table*}[t]
\centering
\begin{minipage}[t]{0.65\textwidth}
\centering
\caption{KPIs for support and balance (mean $\pm$ std).}
\label{tab:kpi-support}
\begin{tabular}{lcccc}
\toprule
Algorithm & SideSup & SurfSup & CoG2D & CoG3D \\
\midrule
O3D-BPP-PCT & 0.222 $\pm$ 0.165 & 0.514 $\pm$ 0.153 & 0.631 $\pm$ 0.195 & 0.628 $\pm$ 0.196 \\
GOPT        & 0.029 $\pm$ \textbf{0.049} & 0.408 $\pm$ 0.122 & 0.527 $\pm$ 0.152 & 0.632 $\pm$ 0.151 \\
Sisyphus    & 0.195 $\pm$ 0.098 & 0.656 $\pm$ \textbf{0.100} & 0.601 $\pm$ 0.136 & 0.581 $\pm$ \underline{0.078} \\
Extreme Point & 0.264 $\pm$ 0.085 & \underline{0.707} $\pm$ 0.152 & 0.752 $\pm$ 0.133 & \underline{0.697} $\pm$ 0.108 \\
MaxRects    & \underline{0.436} $\pm$ 0.096 & 0.441 $\pm$ 0.115 & 0.538 $\pm$ \underline{0.110} & 0.357 $\pm$ 0.124 \\
GENPACK (core) & \textbf{0.561} $\pm$ \underline{0.076} & 0.583 $\pm$ 0.121 & \textbf{0.851} $\pm$ \textbf{0.057} & \textbf{0.748} $\pm$ \textbf{0.071} \\
GENPACK (+PP) & 0.382 $\pm$ 0.099 & \textbf{0.815} $\pm$ \underline{0.114} & \underline{0.803} $\pm$ 0.111 & 0.670 $\pm$ 0.110 \\
\bottomrule
\end{tabular}
\end{minipage}
\hfill
\begin{minipage}[t]{0.32\textwidth}
\centering
\caption{Execution time over 1500 orders.}
\label{tab:runtime}
\begin{tabular}{lrr}
\toprule
Algorithm & Mean (s) & Std (s) \\
\midrule
O3D-BPP-PCT & \underline{0.25} & \underline{0.28} \\
GOPT & 2.80 & 1.23 \\
Sisyphus & 6.00 & 20.86 \\
Extreme Point & 14.29 & 6.69 \\
MaxRects & \textbf{0.02} & \textbf{0.01} \\
GENPACK (core) & 3.84 & 2.15 \\
GENPACK (+PP) & 29.20 & 28.09 \\
\bottomrule
\end{tabular}
\end{minipage}
\\[2pt]
\footnotesize\textit{Note: Best values are in bold; second-best are underlined.}
\end{table*}

\begin{figure*}[t!]
    \centering
    \begin{subfigure}[t]{0.673\textwidth}
        \centering
        \includegraphics[width=\linewidth]{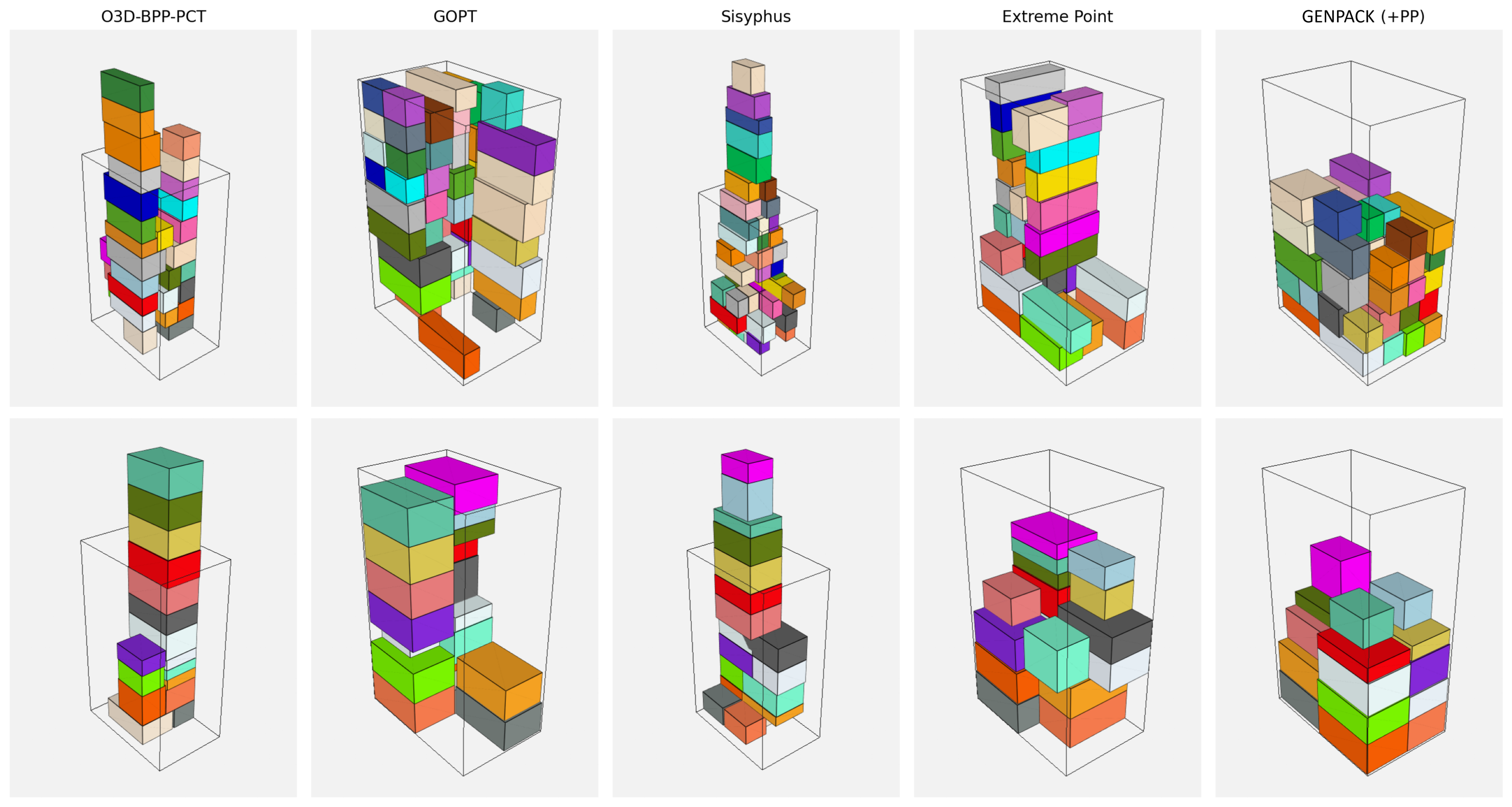}
        \caption{}
        \label{fig:comparison}
    \end{subfigure}
    \hfill
    \begin{subfigure}[t]{0.313\textwidth}
        \centering
        \includegraphics[width=\linewidth]{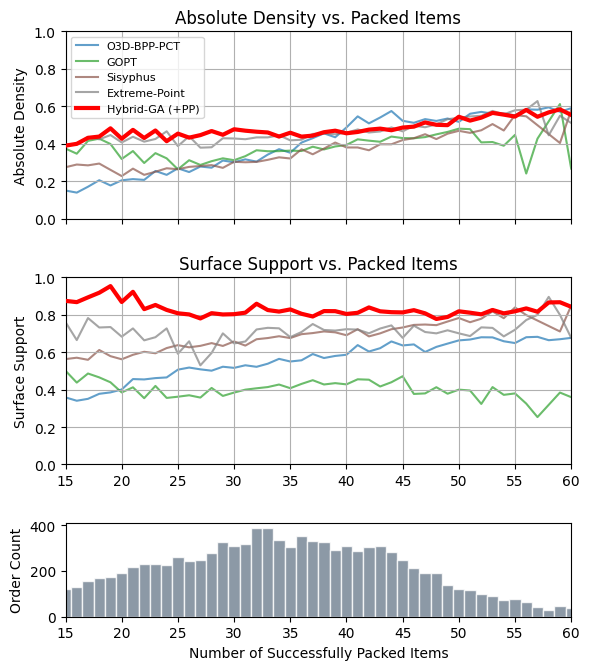}
        \caption{}
        \label{fig:kpi-graphs}
    \end{subfigure}
    \caption{Qualitative (left) and quantitative (right) comparison of packing performance across algorithms on two representative orders from the BED-BPP dataset~\cite{kagerer2023bedbpp}. The left panel illustrates structural differences in packing outcomes, while the right plots absolute density and surface support as functions of packed items, with order-size distribution shown below.}
    \label{fig:comparison-overview}
\end{figure*}

\section{Experimental Results}

\subsection{KPI-Based Analysis}

We first analyze solution quality across seven KPIs, grouped into efficiency/stability and support/balance. Tables~\ref{tab:kpi-density} and~\ref{tab:kpi-support} report scores, with best values highlighted.

\paragraph{Efficiency and Stability.}
Table~\ref{tab:kpi-density} shows that \textit{GENPACK (core)} achieves the highest scores in absolute and relative density as well as height--width ratio, confirming that GA refinement effectively compacts items and reduces vertical instability. It also exhibits the lowest standard deviations across all efficiency metrics, demonstrating superior consistency and robustness across diverse order configurations. In contrast, pure heuristics such as MaxRects underperform in density, while Extreme Point achieves moderately good stability but lags in utilization. The DRL baselines (O3D-BPP-PCT, GOPT) show high variance, indicating less predictable performance.

\paragraph{Support and Balance.}
As shown in Table~\ref{tab:kpi-support}, GA refinement also strengthens balance, with \textit{GENPACK (core)} achieving the best overall center-of-gravity alignment. \textit{GENPACK (+PP)} attains the highest surface support, reflecting its ability to improve contact stability across layers, albeit at the cost of reduced side support, indicating a trade-off between lateral and vertical stability. Beyond mean performance, \textit{GENPACK (core)} demonstrates the lowest variability in center-of-gravity measures and strong consistency, confirming that GA-based optimization yields reliable and stable balance across different problem instances. Learning-based methods (O3D-BPP-PCT, GOPT) achieve moderate performance but remain consistently weaker than GA-based pipelines, exhibiting on average larger variance and thus less predictable solution quality.

In addition to aggregate KPI scores, Figure~\ref{fig:kpi-graphs} shows the evolution of absolute density and surface support as a function of the number of successfully packed items. The histogram (bottom) reveals that order sizes are concentrated in the range of 15--60 items, with the distribution peaking in the 25--35 item range and maintaining substantial representation through 45 items, thus motivating our focused analysis. On the efficiency side (top), O3D-BPP-PCT starts from low density and shows only limited improvement across the range, while GOPT and Sisyphus remain consistently and significantly below the densities of \textit{GENPACK (+PP)} and Extreme Point. The latter attains comparable density trends but remains below \textit{GENPACK (+PP)} on average, which maintains steady improvement and achieves the highest density across this critical range. On the stability side (middle), O3D-BPP-PCT remains relatively weak throughout, GOPT and Sisyphus similarly underperform, and Extreme Point, though more stable, achieves lower surface support than \textit{GENPACK (+PP)}. The latter exhibits both higher and flatter curves, reflecting its ability to preserve surface contact and structural robustness across varying order sizes. These qualitative trends supplement the tabular results in Table~\ref{tab:kpi-support} by showing that GENPACK pipelines not only achieve stronger KPI averages but also maintain robustness across the distribution of order sizes.

Figure~\ref{fig:comparison} shows representative packings. Learning-based baselines (O3D-BPP-PCT, GOPT) often produce tall and irregular stacks with noticeable gaps on this real-world industrial dataset, while heuristic methods (Sisyphus, Extreme Point) yield more compact structures but with vertical instability or uneven support. While Sisyphus attains surface-support values comparable to those of \textit{GENPACK (+PP)}, it frequently produces tall, unstable stacks, showing that numerical support alone does not guarantee feasible industrial packings. In contrast, the GENPACK pipelines generate more balanced and layered configurations, with \textit{GENPACK (+PP)} achieving higher compaction and stronger surface support.

Inference time is crucial for large-scale industrial use, where thousands of orders are processed daily. Table~\ref{tab:runtime} reports statistics for execution times across 1500 orders. While heuristic and learning-based methods achieved sub-second to few-second runtimes, GENPACK variants required longer execution (\textasciitilde 4--29 s per order) due to their multi-stage optimization.

\section{Discussion}

The results highlight a clear trade-off between solution quality and inference time. The KPI-focused analysis shows that the hybrid pipeline consistently outperforms all baselines across density, stability, support, and balance metrics. These improvements translate directly into safer, more robust pallet configurations that reduce handling risks and improve downstream efficiency.

While heuristics and learning-based approaches process orders nearly instantaneously, they do so at the expense of solution quality. In contrast, the GENPACK (+PP) requires longer runtimes, up to 29 seconds per order for the full pipeline. In warehouse contexts, where optimization can be run offline or in parallel, these runtimes can still be considered acceptable given the significant improvements in packing feasibility.

Overall, the results indicate that the primary bottleneck in industrial deployment is not speed alone, but solution quality under operational constraints. Fast heuristics and neural methods can deliver throughput but often fail to guarantee stability or balanced load distribution. By explicitly optimizing multiple KPIs, our method achieves configurations that are both efficient and safe, even if this comes at higher computational cost.

\section{Conclusion and Future Work}

We presented a genetic algorithm (GA) tailored to industrial 3D bin packing and embedded it within a hybrid pipeline combining constructive heuristics and post-processing. The method employs a KPI-guided fitness function addressing utilization, stability, support, and balance; a layer-based chromosome preserving feasibility; and structure-aware operators. These components enable the GA to perform global refinements beyond rule-based placement while remaining compatible with palletization constraints.

Compared with constructive strategies such as MaxRects~\cite{jylanki2010maxrects} and Extreme Point~\cite{crainic2008extreme}, which prioritize speed and initial utilization, the GA provides incremental improvements in balance, stability, and support. Its role is not to generate full packings, but to refine them toward industrially relevant criteria. These improvements come with slightly higher computational cost, making the overall benefit dependent on latency budgets and the chosen KPI weighting.

Current limitations include the use of axis-aligned cuboidal items, proxy-based stability and support measures, and a layer abstraction that may restrict fine-grained placements. Performance also depends on KPI scalarization, which may require tuning for different order mixes or domains.

Future work will explore adaptive and data-driven scalarization for dynamically adjusting KPI weights, as well as extensions to online or uncertain scenarios. We further plan to investigate learning-guided operator control and tighter integration between GA refinement, feasibility checks, and post-processing. Finally, hybridization with other metaheuristics and the incorporation of physical or robotics-aware constraints remain promising directions for improving industrial applicability.

\begin{acks}
This work was supported by Technische Hochschule Augsburg and by the Hightech Agenda Bavaria, funded by the Free State of Bavaria, Germany. The authors also thank their colleagues at Technologietransferzentrum (TTZ) Landsberg for Data Science and Autonomous Systems for their valuable support and discussions throughout this research.
\end{acks}

\bibliographystyle{ACM-Reference-Format}

\begin{thebibliography}{99}

\bibitem{martello2000three}
Silvano Martello, David Pisinger, and Daniele Vigo. 2000. The three-dimensional bin packing problem. \emph{Operations Research} 48, 2 (2000), 256--267.

\bibitem{gehring1997genetic}
Herbert Gehring and Andreas Bortfeldt. 1997.
A genetic algorithm for solving the container loading problem.
\emph{International Transactions in Operational Research} 4, 5--6 (1997), 401--418.
DOI: https://doi.org/10.1111/j.1475-3995.1997.tb00095.x

\bibitem{bortfeldt2001hybrid}
Andreas Bortfeldt and Hermann Gehring. 2001.
A hybrid genetic algorithm for the container loading problem.
\emph{European Journal of Operational Research} 131, 1 (2001), 143--161.

\bibitem{egeblad2009heuristic}
Jens Egeblad and David Pisinger. 2009.
Heuristic approaches for the two- and three-dimensional knapsack packing problem.
\emph{Computers \& Operations Research} 36, 4 (2009), 1026--1049.

\bibitem{bortfeldt2013constraints}
Bortfeldt, Andreas, and Gerhard Wäscher.
\newblock Constraints in container loading -- a state-of-the-art review.
\newblock \emph{European Journal of Operational Research}, 229(1):1--20, 2013.

\bibitem{chazelle1983bottomleft}
Bernard Chazelle. 1983. The bottom-left bin-packing heuristic: an efficient implementation. \emph{IEEE Transactions on Computers} C-32, 8 (Aug. 1983), 697–707. DOI:https://doi.org/10.1109/TC.1983.1676307

\bibitem{crainic2008extreme}
Teodor Gabriel Crainic, Guido Perboli, and Roberto Tadei. 2008.
Extreme point-based heuristics for three-dimensional bin packing.
\emph{INFORMS Journal on Computing} 20, 3 (2008), 368--384.

\bibitem{jylanki2010maxrects}
Jukka Jylänki. 2010. A thousand ways to pack the bin -- A practical approach to two-dimensional rectangle bin packing. Technical Report. Helsinki Institute for Information Technology.

\bibitem{bischoff1990three}
Bischoff, Eberhard E., and Michael D. Marriott. 1990. A comparative evaluation of heuristics for container loading. \emph{European Journal of Operational Research} 44, 2 (1990), 267--276. DOI:https://doi.org/10.1016/0377-2217(90)90362-F

\bibitem{ribeiro2023toolbox}
Luis Ribeiro and Anan Ashrabi Ananno. 2023. A software toolbox for realistic dataset generation for testing online and offline 3D bin packing algorithms. \emph{Processes} 11, 7 (2023), 1909.

\bibitem{jakobs1996genetic}
Stefan Jakobs. 1996. On genetic algorithms for the packing of polygons. \emph{European Journal of Operational Research} 88, 1 (1996), 165--181.

\bibitem{wu2010three}
Wenbin Wu, Lixin Li, Kien Goh, and Roberto De Souza. 2010. A hybrid genetic algorithm for the three-dimensional bin packing problem. \emph{Applied Mathematics and Computation} 216, 9 (2010), 2703--2713.

\bibitem{kang2012hybrid}
Kyungdaw Kang, Ilkyeong Moon, and Hongfeng Wang. 2012. 
A hybrid genetic algorithm with a new packing strategy for the three-dimensional bin packing problem. 
\emph{Applied Mathematics and Computation} 219, 3 (Oct. 2012), 1287--1299. DOI:https://doi.org/10.1016/j.amc.2012.07.036

\bibitem{ananno2024multiheuristic}
Anan Ashrabi Ananno and Luis Ribeiro. 2024. 
A multi-heuristic algorithm for multi-container 3-D bin packing problem optimization using real-world constraints. 
\emph{IEEE Access} 12 (2024), 42105--42130. 
DOI:~https://doi.org/10.1109/ACCESS.2024.3378063.

\bibitem{kagerer2023bedbpp}
Florian Kagerer, Maximilian Beinhofer, Stefan Stricker, and Andreas Nüchter. 2023. BED-BPP: Benchmarking dataset for robotic bin packing problems. \emph{The International Journal of Robotics Research} 42, 11 (2023), 1007--1014. DOI:https://doi.org/10.1177/02783649231193048

\bibitem{ceschia2013metaheuristic}
Sara Ceschia and Andrea Schaerf. 2013. Local search techniques for a routing-packing problem. \emph{Computers \& Industrial Engineering} 66, 4 (2013), 1138--1149.

\bibitem{eiben1999parameter}
Agoston E. Eiben, Robert Hinterding, and Zbigniew Michalewicz. 1999. 
Parameter control in evolutionary algorithms. 
\emph{IEEE Transactions on Evolutionary Computation} 3, 2 (1999), 124--141.

\bibitem{zhao2021constrained}
Hang Zhao, Qijin She, Chenyang Zhu, Yin Yang, and Kai Xu. 2021. Online 3D bin packing with constrained deep reinforcement learning. \emph{Proceedings of the AAAI Conference on Artificial Intelligence} 35, 1 (2021), 741--749. DOI:https://doi.org/10.1609/aaai.v35i1.16155

\bibitem{zhao2022pct}
Hang Zhao, Yang Yu, and Kai Xu. 2022. Learning efficient online 3D bin packing on packing configuration trees. In \emph{International Conference on Learning Representations (ICLR 2022)}.

\bibitem{xiong2024gopt}
Heng Xiong, Changrong Guo, Jian Peng, Kai Ding, Wenjie Chen, Xuchong Qiu, Long Bai, and Jianfeng Xu. 2024. 
GOPT: Generalizable online 3D bin packing via transformer-based deep reinforcement learning. 
\emph{IEEE Robotics and Automation Letters} 9, 11 (Nov. 2024), 10335--10342. 
DOI:~https://doi.org/10.1109/LRA.2024.3468161.

\bibitem{que2023transformer}
Quanqing Que, Fang Yang, and Defu Zhang. 2023. Solving 3D packing problem using Transformer network and reinforcement learning. \emph{Expert Systems with Applications} 214 (2023), 119153. DOI:https://doi.org/10.1016/j.eswa.2022.119153

\bibitem{elhedhli2019three}
Samir Elhedhli, Fatma Gzara, and Burak Yildiz. 2019. Three-dimensional bin packing and mixed-case palletization. \emph{INFORMS Journal on Optimization} 1, 4 (2019), 323--352.

\bibitem{ramos2018new}
António G. Ramos, Joaquim J. Júdice, and José F. Oliveira. 2018. A new load balance methodology for container loading problem in road transportation. \emph{European Journal of Operational Research} 266, 3 (2018), 1140--1152.

\bibitem{fanslau2010tree}
Thomas Fanslau and Andreas Bortfeldt. 2010. A tree search algorithm for solving the container loading problem. \emph{INFORMS Journal on Computing} 22, 2 (2010), 222--235.

\bibitem{pantoja2024comprehensive}
Gabriel Pantoja-Benavides, David Giraldo, Andrés Montes, Diego Ruiz, Carlos Franco, and Jaime Álvarez-Gallego. 2024. Comprehensive review of robotized freight packing: From theoretical foundations to practical implementations. \emph{Logistics} 8, 3 (2024), 69.

\bibitem{wu2023machine}
Wenjie Wu, Changjun Fan, Jincai Huang, Zhong Liu, and Junchi Yan. 2023. Machine learning for the multi-dimensional bin packing problem: literature review and empirical evaluation. \emph{arXiv} abs/2312.08103 (2023). DOI:https://doi.org/10.48550/arXiv.2312.08103

\bibitem{demisse2012mixed}
Girum Demisse, Razvan Mihalyi, Billy Okal, Dev Poudel, Johannes Schauer, and Andreas Nüchter. 2012. Mixed palletizing and task completion for virtual warehouses. In \emph{Proceedings of the Virtual Manufacturing and Automation Competition (VMAC) Workshop at the IEEE International Conference on Robotics and Automation (ICRA), Vol. 16}. IEEE, 2012.

\end{thebibliography}

\newcommand{\leni}{\tilde l_i(\theta_i)}
\newcommand{\widi}{\tilde w_i(\theta_i)}
\newcommand{\lenj}{\tilde l_j(\theta_j)}
\newcommand{\widj}{\tilde w_j(\theta_j)}

\appendix

\section{Post-Processing}
\label{app:postprocessing}

The post-processing procedure in Algorithm~\ref{alg:postprocessing} can be expressed as a sequence of constrained optimization steps applied to the GA-refined packing $S^\star$. Each stage operates under the same geometric and physical feasibility rules defined below.

\subsection{Notation and Feasibility Conditions}

For completeness, we restate the main notation used in the paper. Each item~$i$ has dimensions $(l_i, w_i, h_i)$ and a placement $(x_i, y_i, z_i, \theta_i)$ within a pallet of size $(L, W, H)$. The rotation angle $\theta_i \in \{0,90^\circ\}$ swaps the item's width and length when rotated. An item is considered \emph{feasible} if it satisfies the following conditions:

\begin{enumerate}
  \item \textbf{Containment:} the item lies completely within the pallet bounds.
  \item \textbf{Non-overlap:} it does not intersect any other item in three dimensions.
  \item \textbf{Support:} its base area is sufficiently supported by underlying items, with a support ratio $\rho_{\mathrm{support}}(i)$ above the threshold~$\tau_s$.
\end{enumerate}

The support threshold varies by stage, typically $0.75$ during compaction and at least $0.5$ in final validation. The feasible placements of item~$i$ form the set~$\mathcal{F}_i$, defined by these three conditions.

\subsection{Stage 1: Directional Compaction}

Directional compaction reduces voids by shifting items toward the pallet origin while maintaining feasibility. For each layer~$\mathcal{L}_j$, the new horizontal coordinates are obtained as
\begin{equation}
\begin{split}
(x_i', y_i')
&= \arg\min_{\{(x_i, y_i)\}}
    \sum_{i \in \mathcal{L}_j} (x_i + y_i) \\
&\quad\text{s.t.}\quad
    (x_i, y_i, z_j, \theta_i) \in \mathcal{F}_i,
\end{split}
\label{eq:compaction}
\end{equation}
where $z_j$ denotes the common base height of layer~$j$, shared by all items $i \in \mathcal{L}_j$. The updated positions $(x_i', y_i')$ replace the previous coordinates, producing a compacted layout that minimizes total offset $\sum_i (x_i + y_i)$ while preserving feasibility. Discrete moves $(\Delta x, \Delta y)\!\in\!\{(-s,0),(0,-s),(-s,-s)\}$ with step sizes $s\!\in\!\{25,10,5,1\}$ are tested, and a move is accepted only if it reduces $(x_i + y_i)$ and maintains feasibility.

\subsection{Stage 2: Fallback Placement}

Unplaced items are recovered through a grid-based search. Candidate positions are generated as
\[
\mathcal{C}_i =
\{(x, y, z, \theta)\,|\,x\in X,\ y\in Y,\ z\in\mathcal{Z},\ \theta\in\{0,90^\circ\}\},
\]
where $X$ and $Y$ denote discrete grids of feasible horizontal coordinates,
\[
X = \{0, \Delta, 2\Delta, \dots, L - l_i\}, \qquad
Y = \{0, \Delta, 2\Delta, \dots, W - w_i\},
\]
with $\Delta$ being the grid step size that controls the spatial resolution of the search. The upper limits $L - l_i$ and $W - w_i$ ensure that each item remains fully contained within the pallet boundaries. The set $\mathcal{Z}$ represents candidate stacking heights, including the ground level and the top surfaces of previously placed items that can provide sufficient support, i.e.
\[
\mathcal{Z} = \{0\} \cup \{\, z_k + h_k \mid k 
\text{ is a feasible supporting item}\,\}.
\]
Thus, $\mathcal{C}_i$ enumerates all discretized placement candidates $(x, y, z, \theta)$ for item~$i$, considering both orientations $\theta \in \{0, 90^\circ\}$. Among feasible candidates, the selected position minimizes height and offset:
\begin{equation}
(x_i, y_i, z_i, \theta_i) =
\arg\min_{(x', y', z', \theta')\in\mathcal{C}_i\cap\mathcal{F}_i}
[z',\; x'+y']_{\mathrm{lex}},
\label{eq:fallback}
\end{equation}
where lexicographic order prioritizes lower $z'$ and then smaller $(x'+y')$. If no feasible position exists, the grid step~$\Delta$ is halved and the search is repeated for up to two refinement rounds.

\subsection{Stage 3: Validation}

The final validation step removes any infeasible or unsupported items. Each item is assigned an indicator
\[
\chi_i =
\begin{cases}
1,& (x_i, y_i, z_i, \theta_i)\in\mathcal{F}_i,\\
0,& \text{otherwise},
\end{cases}
\]
and the validated configuration is
\begin{equation}
S^{\star}_{\mathrm{pp}}
= \{\, i \in S^\star \mid \chi_i = 1 \,\}.
\label{eq:validated}
\end{equation}
The resulting post-processed solution $S^{\star}_{\mathrm{pp}}$ satisfies all geometric and physical constraints and forms a stable pallet configuration.

\section{Additional KPI trends}
\label{sec:addkpis}

\begin{figure}[t]
    \centering
    \includegraphics[width=0.8\linewidth]{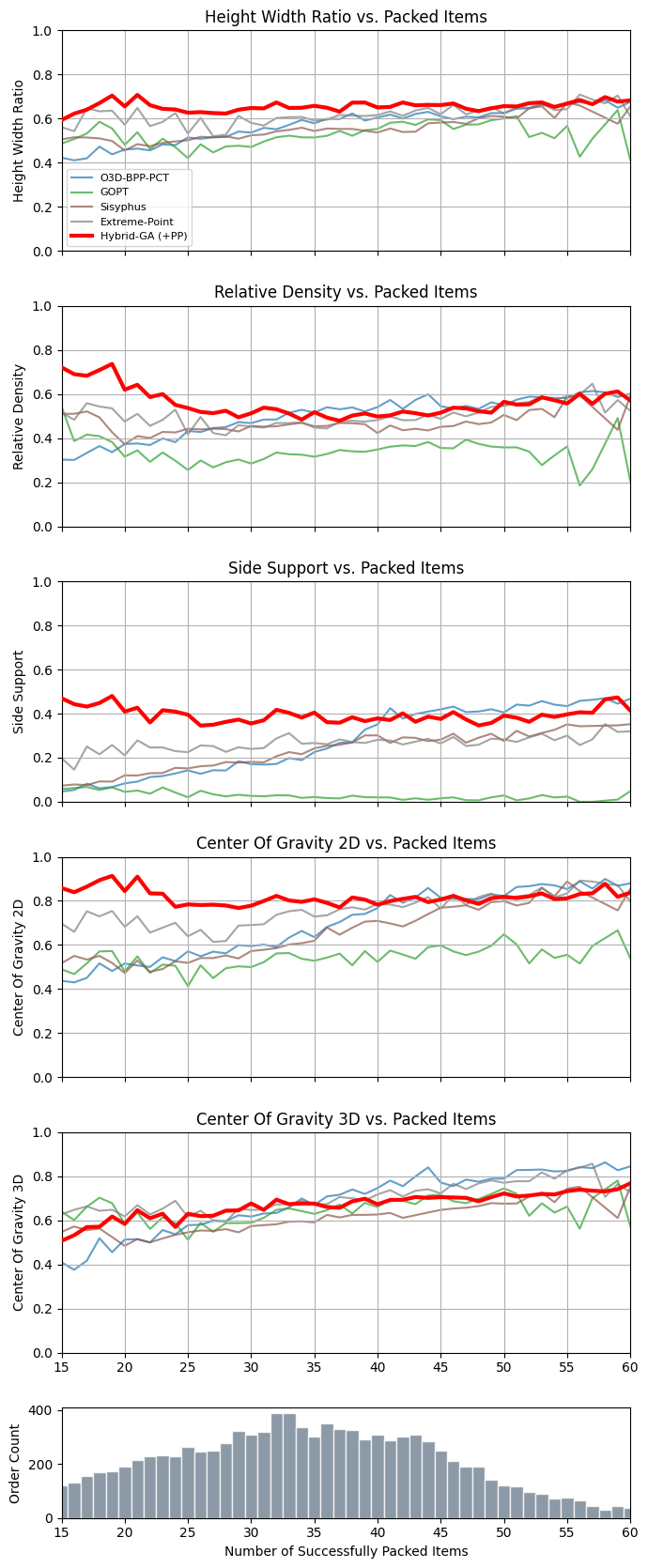}
    \caption{Supplementary analysis of additional KPIs, including height-width ratio, relative density, side support, and center-of-gravity balance as functions of packed items, with order-size distribution shown below.}
    \label{fig:kpi-appendix}
\end{figure}

\paragraph{Relative Density and Height--Width Ratio.}
Figure~\ref{fig:kpi-appendix} (top) compares \textit{Relative Density} and \textit{Height--Width Ratio} across methods. \textit{GENPACK (+PP)} achieves consistently high relative density and remains at or above the baselines across most of the range of packed items. O3D-BPP-PCT starts lower but steadily improves, approaching GENPACK (+PP) at higher item counts; Extreme Point remains mid-range, while Sisyphus and GOPT fluctuate and dip more frequently. For \textit{Height--Width Ratio}, GENPACK (+PP) and Extreme Point maintain comparatively higher and more stable ratios, whereas O3D-BPP-PCT, Sisyphus, and GOPT exhibit greater variance, indicating less consistent vertical compactness.

\paragraph{Support and Balance.}
In the lower plots of Figure~\ref{fig:kpi-appendix}, \textit{Side Support} generally increases with the number of packed items. GENPACK (+PP) maintains consistent support throughout, whereas O3D-BPP-PCT struggles in the lower-item range but recovers to achieve the strongest support at higher item counts. Extreme Point follows closely behind, while GOPT remains lowest on average. \textit{CoG2D} values for all methods cluster near the center as item count increases; GENPACK (+PP) exhibits slightly tighter dispersion, indicating steadier horizontal balance. \textit{CoG3D} shows a similar convergence trend at higher counts, with early-bin variability more pronounced for Sisyphus and GOPT. Overall, GENPACK (+PP) is competitive or leading in both compactness and stability, while O3D-BPP-PCT improves markedly as instance size grows.

\end{document}